\documentclass[11pt,a4paper]{article}
\usepackage[hyperref]{acl2019}
\usepackage{times}
\usepackage{latexsym}

\usepackage{url}

\usepackage{amsmath}
\usepackage{float}
\usepackage{graphicx}
\usepackage{longtable}
\usepackage[normalem]{ulem}

\usepackage{amsmath,amssymb,amsthm}
\usepackage{subcaption,siunitx,booktabs}
\usepackage{makecell}
\usepackage{arydshln}
\usepackage{xspace}
\usepackage{soul}
\usepackage{color}
\usepackage{multirow, booktabs}

\aclfinalcopy % Uncomment this line for the final submission
\def\aclpaperid{518} %  Enter the acl Paper ID here

\title{Improving Low-Resource Cross-lingual Document Retrieval by Reranking with Deep Bilingual Representations}
\author{Rui Zhang \quad\quad Caitlin Westerfield \quad\quad Sungrok Shim \\ {\bf Garrett Bingham \quad Alexander Fabbri \quad William Hu \quad Neha Verma \quad Dragomir Radev} \\ Department of Computer Science, Yale University \\ {\tt\{r.zhang, dragomir.radev\}@yale.edu} }
\date{}

\begin{document}
\maketitle
\begin{abstract}
In this paper, we propose to boost low-resource cross-lingual document retrieval performance with deep bilingual query-document representations.
We match queries and documents in both source and target languages with four components, each of which is implemented as a term interaction-based deep neural network with cross-lingual word embeddings as input.
By including query likelihood scores as extra features, our model effectively learns to rerank the retrieved documents by using a small number of relevance labels for low-resource language pairs.
Due to the shared cross-lingual word embedding space, the model can also be directly applied to another language pair without any training label.
Experimental results on the \textsc{Material} dataset show that our model outperforms the competitive translation-based baselines on English-Swahili, English-Tagalog, and English-Somali cross-lingual information retrieval tasks.
\end{abstract}

% \vspace{-3mm}
\section{Introduction}
% \vspace{-2mm}
% Introduce CLIR and why it is important
Cross-lingual relevance ranking, or Cross-Lingual Information Retrieval (CLIR), is the task of ranking foreign documents against a user query \cite{hull1996querying,ballesteros1996dictionary,oard1997document,darwish2003probabilistic}.
As multilingual documents are more accessible, CLIR is increasingly more important whenever the relevant information is in other languages.

Traditional CLIR systems consist of two components: machine translation and monolingual information retrieval.
Based on the translation direction, it can be further categorized into the document translation and the query translation approaches \cite{nie2010cross}.
In both cases, we first solve the translation problem, and the task is transformed to the monolingual setting.
However, while conceptually simple, the performance of this modular approach is fundamentally limited by the quality of machine translation.

\begin{figure}[t!]
  \centering
  \includegraphics[width=\columnwidth]{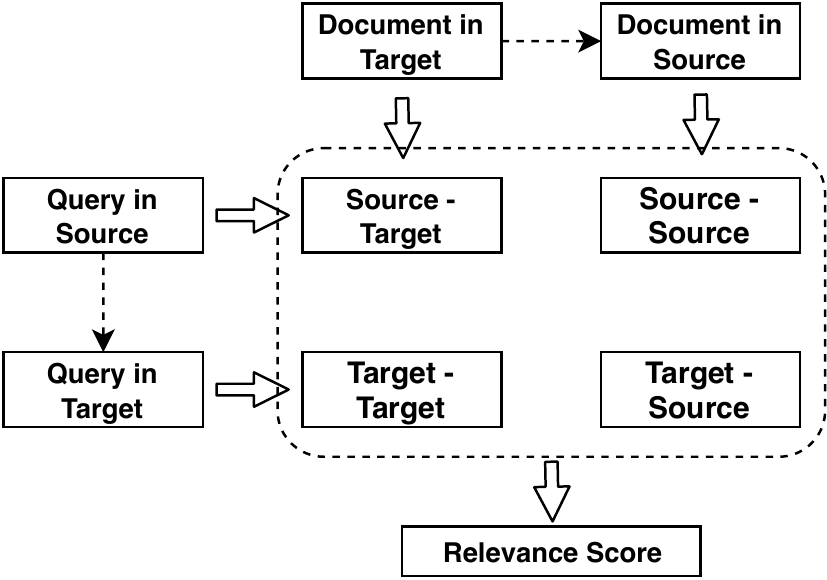}
  \caption{Cross-lingual Relevance Ranking with Bilingual Query and Document Representation.}
%   \vspace{-5mm}
  \label{fig:pipeline}
\end{figure}

% Neural Monolingual IR / Learning to rank
Recently, many deep neural IR models have shown promising results on monolingual data sets \cite{huang2013learning,guo2016deep,pang2016study,mitra2016dual,mitra2017learning,xiong2017end,hui2017pacrr,hui2018co,mcdonald2018deep}.
They learn a scoring function directly from the relevance label of query-document pairs.
However, it is not clear how to use them when documents and queries are not in the same language.
Furthermore, those deep neural networks need a large amount of training data.
This is expensive to get for low-resource language pairs in our cross-lingual case.

\begin{figure*}[ht!]
  \centering
  \begin{subfigure}{.93\textwidth}
      \centering
      \includegraphics[trim={0 0mm 0 0mm},clip,width=.93\textwidth]{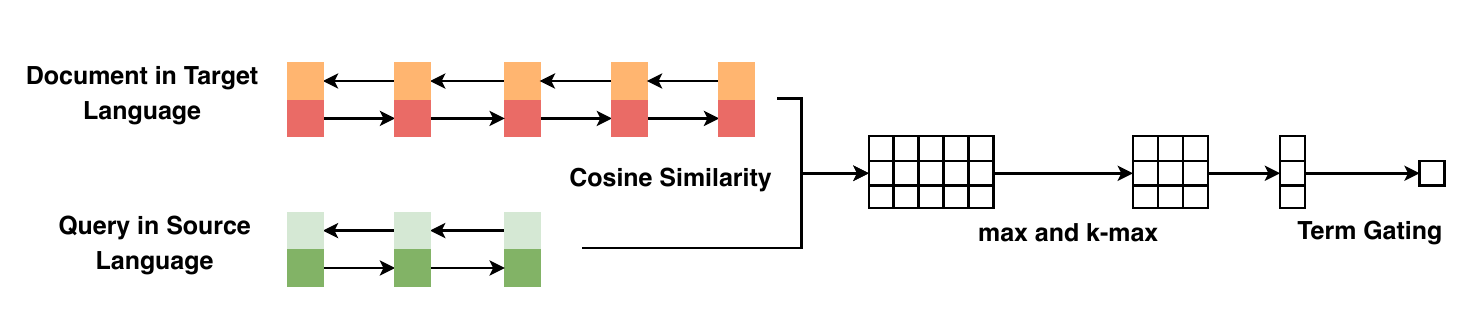}
    %   \vspace{-1mm}
      \caption{Bilingual POSIT-DRMM. The colored box represents hidden states in bidirectional LSTMs.}
    %   \vspace{-1mm}
      \label{fig:posit-drmm}
  \end{subfigure}
  \begin{subfigure}{\textwidth}
      \centering
      \includegraphics[trim={0 0 0 3mm},clip,width=\textwidth]{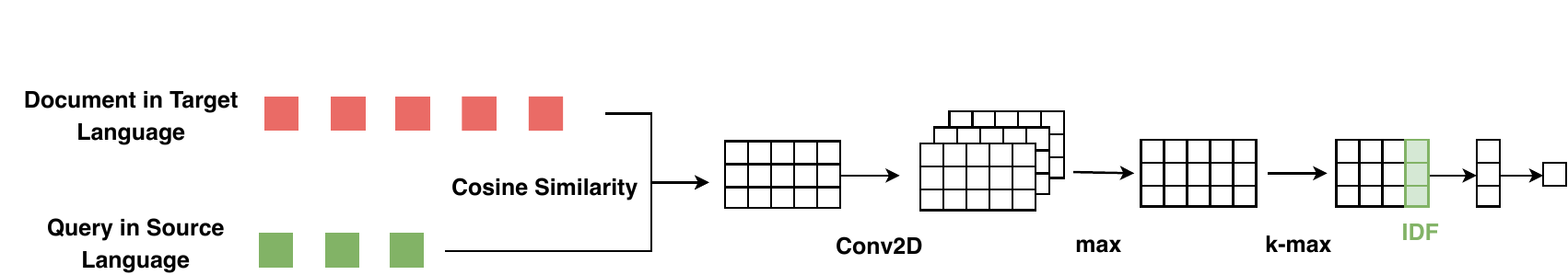}
      \vspace{0.3mm}
      \caption{Bilingual PACRR-DRMM. The colored box represents cross-lingual word embeddings. Bilingual PACRR is the same except it uses a single MLP at the final stage.}
      \label{fig:pacrr-drmm}
  \end{subfigure}
\caption{Model architecture. We only show the component of the source query with the target document.}
% \vspace{-3mm}
\end{figure*}

In this paper, we propose a cross-lingual deep relevance ranking architecture based on a bilingual view of queries and documents.
As shown in Figure \ref{fig:pipeline}, our model first translates queries and documents and then uses four components to match them in both the source and target language.
Each component is implemented as a deep neural network, and the final relevance score combines all components which are jointly trained given the relevance label.
We implement this based on state-of-the-art term interaction models because they enable us to make use of cross-lingual embeddings to explicitly encode terms of queries and documents even if they are in different languages.
To deal with the small amount of training data, we first perform query likelihood retrieval and include the score as an extra feature in our model.
In this way, the model effectively learns to rerank from a small number of relevance labels.
Furthermore, since the word embeddings are aligned in the same space, our model can directly transfer to another language pair with no additional training data.

We evaluate our model on the \textsc{Material} CLIR dataset with three language pairs including English to Swahili, English to Tagalog, and English to Somali.
Experimental results demonstrate that our model outperforms other translation-based query likelihood retrieval and monolingual deep relevance ranking approaches.

% \vspace{-2mm}
\section{Our Method}
% \vspace{-2mm}
\label{sec:method}
In cross-lingual document retrieval, given a user query in the source language $Q$ and a document in the target language $D$,  the system computes a relevance score $s(Q,D)$.
As shown in Figure \ref{fig:pipeline}, our model first translates the document as $\hat{D}$ or the query as $\hat{Q}$, and then it uses four separate components to match: (1) source query with target document, (2) source query with source document, (3) target query with source document, (4) target query with target document.
The final relevance score combines all components:
\resizebox{\columnwidth}{!}{$s(Q,D) = s(Q,D) + s(Q,\hat{D}) + s(\hat{Q},\hat{D}) + s(\hat{Q},D)$}

To implement each component, we extend three state-of-the-art \textit{term interaction models}: PACRR (Position-Aware Convolutional Recurrent Relevance Matching) proposed by \newcite{hui2017pacrr}, POSIT-DRMM (POoled SImilariTy DRMM) and PACRR-DRMM proposed by \newcite{mcdonald2018deep}.
In term interaction models, each query term is scored to a document's terms from the interaction encodings, and scores for different query terms are aggregated to produce the query-document relevance score.
\vspace{-2mm}
\subsection{Bilingual POSIT-DRMM}
\vspace{-1mm}
This model is illustrated in Figure \ref{fig:posit-drmm}.
We first use bidirectional LSTMs \cite{hochreiter1997long} to produce the context-sensitive encoding of each query and document term.
We also add residual connection to combine the pre-trained term embedding and the LSTM hidden states.
For the source query and document term, we can use the pre-trained word embedding in the source language.
For the target query and document term, we first align the pre-trained embedding in the target language to the source language and then use this cross-lingual word embedding as the input to LSTM.
Thereafter, we produce the document-aware query term encoding by applying max pooling and $k$-max pooling over the cosine similarity matrix of query and document terms.
We then use an MLP to produce term scores, and the relevance score is a weighted sum over all terms in the query with a term gating mechanism.
\begin{table*}[ht!]
\centering
\resizebox{.8\textwidth}{!}{
\begin{tabular}{cccccccccc}
\Xhline{4\arrayrulewidth}
                                & \multicolumn{3}{c}{EN-\textgreater{}SW} & \multicolumn{3}{c}{EN-\textgreater{}TL} & \multicolumn{3}{c}{EN-\textgreater{}SO} \\ \hline
\# Document                     & \multicolumn{3}{c}{813}                 & \multicolumn{3}{c}{844}     & \multicolumn{3}{c}{695}  \\
\# Document Token (Min/Avg/Max) & \multicolumn{3}{c}{34/341/1724}         & \multicolumn{3}{c}{32/404/2501}  & \multicolumn{3}{c}{69/370/2671}       \\ \hline
Query Set                       & Q1          & Q2          & Q3          & Q1          & Q2          & Q3    & \multicolumn{3}{c}{Q1}      \\ \hline
\# Query                        & 300         & 400         & 600         & 300         & 400         & 600   & \multicolumn{3}{c}{300}       \\
\# Relevant Pairs               & 411         & 489         & 828         & 236         & 576         & 1018 & \multicolumn{3}{c}{496} \\
\Xhline{4\arrayrulewidth}
\end{tabular}
}
\caption{The MATERIAL dataset statistics.
For SW and TL, we use the ANALYSIS document set with Q1 for training, Q2 for dev, and Q3 for test.
For transfer learning to SO, we use the DEV document set with Q1.
Q1 contains open queries where performers can conduct any automatic or manual exploration while Q2 and Q3 are closed queries where results must be generated with fully automatic systems with no human in the loop.}
\label{tab:data}
\vspace{-5mm}
\end{table*}

\subsection{Bilingual PACRR and Bilingual PACRR-DRMM}
These models are shown in Figure \ref{fig:pacrr-drmm}.
We first align the word embeddings in the target language to the source language and build a query-document similarity matrix that encodes the similarity between the query and document term.
Depending on the query language and document language, we construct four matrices, $\text{SIM}_{Q,D}$, $\text{SIM}_{Q, \hat{D}}$, $\text{SIM}_{\hat{Q}, \hat{D}}$, $\text{SIM}_{\hat{Q}, D}$, for each of the four components.
Then, we use convolutional neural networks over the similarity matrix to extract $n$-gram matching features.
We then use max-pooling and $k$-max-pooling to produce the feature matrix where each row is a document-aware encoding of a query term.
The final step computes the relevance score: Bilingual PACRR uses an MLP on the whole feature matrix to get the relevance score, while Bilingual PACRR-DRMM first uses an MLP on individual rows to get query term scores and then use a second layer to combine them.

% \vspace{-3mm}
\section{Related Work}
% \vspace{-2mm}
\label{sec:related_work}
\textbf{Cross-lingual Information Retrieval}.
Traditional CLIR approaches include document translation and query translation, and more research efforts are on the latter \cite{oard1997document,oard1998comparative,mccarley1999should,franz1999ad}.
Early methods use the dictionary to translate the user query \cite{hull1996querying,ballesteros1996dictionary,pirkola1998effects}.
Other methods include the single best SMT query translation \cite{chin2014cross} and the weighted SMT translation alternatives known as the probabilistic structured query (PSQ) \cite{darwish2003probabilistic,ture2012combining}.
Recently, \newcite{bai2010learning} and \newcite{sokolov2013boosting} propose methods to learn the sparse query-document associations from supervised ranking signals on cross-lingual Wikipedia and patent data, respectively.
Furthermore, \newcite{vulic2015monolingual} and \newcite{litschko2018unsupervised} use cross-lingual word embeddings to represent both queries and documents as vectors and perform IR by computing the cosine similarity.
\newcite{schamoni2014learning} and \newcite{sasaki2018cross} also use an automatic process to build CLIR datasets from Wikipeida articles.\\
\textbf{Neural Learning to Rank}.
Most of neural learning to rank models can be categorized in two groups: representation based \cite{huang2013learning,shen2014learning} and interaction based \cite{pang2016study,guo2016deep,hui2017pacrr,xiong2017end,mcdonald2018deep}.
The former builds representations of query and documents independently, and the matching is performed at the final stage.
The latter explicitly encodes the interaction between terms to direct capture word-level interaction patterns.
For example, the DRMM \cite{guo2016deep} first compares the term embeddings of each pair of terms within the query and the document and then generates fixed-length matching histograms.

% \vspace{-2mm}
\section{Experiments}
% \vspace{-2mm}
\label{sec:experiment}
\begin{table*}[t!]
\resizebox{\textwidth}{!}{
\begin{tabular}{lcccccccc}
\Xhline{4\arrayrulewidth}
                     & \multicolumn{4}{c}{EN-\textgreater{}SW}                                                                  & \multicolumn{4}{c}{EN-\textgreater{}TL}                                                                     \\
                     & MAP                  & P@20 & NDCG@20 & AQWV & MAP & P@20 & NDCG@20 & AQWV \\ \hline
\multicolumn{9}{l}{Query Translation and Document Translation with Indri}        \\ \hline
Dictionary-Based Query Translation (DBQT)                 & 20.93 & 4.86 & 28.65 & 6.50 & 20.01 & 5.42 & 27.01 & 5.93 \\
Probabilistic Structured Query (PSQ)                  & 27.16 & 5.81 & 36.03 & 12.56 & 35.20 & 8.18 & 44.04 & 19.81 \\
Statistical MT (SMT)                  & 26.30 & 5.28 & 34.60 & 13.77 & 37.31 & 8.77 & 46.77 & 21.90 \\
Neural MT (NMT)                  & 26.54 & 5.26 & 34.83 & 15.70 & 33.83 & 8.20 & 43.17 & 18.56 \\ \hline
\multicolumn{9}{l}{Deep Relevance Ranking}         \\ \hline
PACRR                & 24.69 & 5.24 & 32.85 & 11.73 & 32.53 & 8.42 & 41.75 & 17.48\\
PACRR-DRMM           & 22.15 & 5.14 & 30.28 & 8.50 & 32.59 & 8.60 & 42.17 & 16.59 \\
POSIT-DRMM           & 23.91 & 6.04 & 33.83 & 12.06 & 25.16 & 8.15 & 34.80 & 9.28 \\
\hline
\multicolumn{9}{l}{Deep Relevance Ranking with Extra Features in Section \ref{sec:extra_feature}}  \\ \hline
PACRR                & 27.03 & 5.34 & 35.36 & 14.18 & 41.43 & 8.98 & 49.96 & 27.46 \\
PACRR-DRMM           & 25.46 & 5.50 & 34.15 & 12.18 & 35.61 & 8.69 & 45.34 & 22.70 \\
POSIT-DRMM           & 26.10 & 5.26 & 34.27 & 14.11 & 39.35 & 9.24 & 48.41 & 25.01 \\ \hline
\multicolumn{9}{l}{Ours with Extra Features in Section \ref{sec:extra_feature}: In-Language Training}                 \\ \hline
Bilingual PACRR      & 29.64 & 5.75 & 38.27 & 17.87 & 43.02 & 9.63 & 52.27 & 29.12 \\
Bilingual PACRR-DRMM & 26.15 & 5.84 & 35.54 & 12.92 & 38.29 & 9.21 & 47.60 & 22.94 \\
Bilingual POSIT-DRMM & \textbf{30.13} & \textbf{6.28} & \textbf{39.68} & \textbf{18.69} & \textbf{43.67} & \textbf{9.73} & \textbf{52.80} & \textbf{29.12} \\
Bilingual POSIT-DRMM (3-model ensemble) & \textbf{31.60} & \textbf{6.37} & \textbf{41.25} & \textbf{20.19} & \textbf{45.35} & \textbf{9.84} & \textbf{54.26} & \textbf{31.08} \\
\Xhline{4\arrayrulewidth}
\end{tabular}
}
\caption{Test set result on English to Swahili and English to Tagalog. 
We report the TREC ad-hoc retrieval evaluation metrics (MAP, P@20, NDCG@20) and the Actual Query Weighted Value (AQWV).
}
\label{tab:result}
\end{table*}

\begin{table}[t!]
\centering
\resizebox{\columnwidth}{!}{
\begin{tabular}{lccc}
\Xhline{4\arrayrulewidth}
\multicolumn{4}{c}{Train: EN-\textgreater{}SW + EN-\textgreater{}TL, Test: EN-\textgreater{}SO} \\
                     & MAP                  & P@20  & AQWV \\ \hline
PSQ                  &  17.52 & 5.45 & 2.35\\
SMT                  &  19.04 & 6.12 & 4.62\\
\hline
Bilingual POSIT-DRMM &  \textbf{20.58} & \textbf{6.51} & \textbf{5.71}\\
\quad +3-model ensemble & \textbf{21.25} & \textbf{6.68} & \textbf{5.89}\\
\Xhline{4\arrayrulewidth}
\end{tabular}
}
\caption{Zero-shot transfer learning on English to Somali test set.}
\label{tab:result_so}
\end{table}
\textbf{Training and Inference}.
We first use the Indri\footnote{\url{www.lemurproject.org/indri.php}} system which uses query likelihood with Dirichlet Smoothing \cite{zhai2004study} to pre-select the documents from the collection.
To build the training dataset, for each positive example in the returned list, we randomly sample one negative example from the documents returned by Indri.
The model is then trained with a binary cross-entropy loss.
On validation or testing set, we use our prediction scores to rerank the documents returned by Indri.\\
\textbf{Extra Features}.
\label{sec:extra_feature}
Following the previous work \cite{severyn2015learning,mohan2017deep,mcdonald2018deep}, we compute the final relevance score by a linear model to combine the model output with the following set of extra features: (1) the Indri score with the language modeling approach to information retrieval. (2) the percentage of query terms with an exact match in the document, including the regular percentage and IDF weighted percentage. (3) the percentage of query term bigrams matches in the document.\\
\textbf{Cross-lingual Word Embeddings}.
We apply the supervised iterative Procrustes approach \cite{xing2015normalized,conneau2018word} to align two pretrained mono-lingual fastText \cite{bojanowski2016enriching} word embeddings using the MUSE implementation\footnote{\url{github.com/facebookresearch/MUSE}}.
To build the bilingual dictionary, we use the translation pages of Wiktionary\footnote{\url{https://www.wiktionary.org/}}.
For Swahili, we build a training dictionary for 5301 words and a testing dictionary for 1326 words.
For Tagalog, the training dictionary and testing dictionary contains 7088 and 1773 words, respectively.
For Somali, the corresponding number is 7633 and 1909.
We then learn the cross-lingual word embeddings from Swahili to English, from Tagalog to English, and from Somali to English.
Therefore, all three languages are in the same word embedding space.\\
\textbf{Data Sets and Evaluation Metrics}.
Our experiments are evaluated on the MATERIAL\footnote{\url{www.iarpa.gov/index.php/research-programs/material}} program as summarized in Table \ref{tab:data}.
It consists of three language pairs with English queries on Swahili (EN-\textgreater{}SW), Tagalog (EN-\textgreater{}TL), Somali documents (EN-\textgreater{}SO).

We use the TREC ad-hoc retrieval evaluation script\footnote{\url{https://trec.nist.gov/trec_eval/}} to compute Precision@20, Mean Average Precision (MAP), Normalized Discounted Cumulative Gain@20 (NDCG@20).
We also report the Actual Query Weighted Value (AQWV) \cite{aqwv}, a set-based metric with penalty for both missing relevant and returning irrelevant documents.
We use $\beta=40.0$ and find the best global fixed cutoff over all queries.\\
\textbf{Baselines}.
For traditional CLIR approaches, we use query translation and document translation with the Indri system.
For query translation, we use Dictionary-Based Query Translation (DBQT) and Probabilistic Structured Query (PSQ).
For document translation, we use Statistical Machine Translation (SMT) and Neural Machine Translation (NMT).
For SMT, we use the moses system \cite{koehn2007moses} with word alignments using mGiza and 5-gram KenLM language model \cite{heafield2011kenlm}.
For NMT, we use sequence-to-sequence model with attention \cite{bahdanau2015neural,barone2017deep} implemented in Marian \cite{mariannmt}.

For deep relevance ranking baselines, we investigate recent state-of-the-art models including PACRR, PACRR-DRMM, and POSIT-DRMM.
These models and our methods all use an SMT-based document translation as input.\\
\textbf{Implementation Details}.
For POSIT-DRMM and Bilingual POSIT-DRMM, we use the $k$-max-pooling with $k=5$ and 0.3 dropout of the BiLSTM output.
For PACRR, PACRR-DRMM and their bilingual counterparts, we use convolutional filter sizes with [1,2,3], and each filter size has 32 filters.
We use $k=2$ in the $k$-max-pooling.
The loss function is minimized using the Adam optimizer \cite{kingma2014adam} with the training batch size as 32.
We monitor the MAP performance on the development set after each epoch of training to select the model which is used on the test data.

% \vspace{-1mm}
\subsection{Results and Discussion}
% \vspace{-1mm}
\label{sec:result}
Table \ref{tab:result} shows the result on EN-\textgreater{}SW and EN-\textgreater{}TL where we train and test on the same language pair.\\
\textbf{Performance of Baselines}.
For query translation, PSQ is better than DBQT because PSQ uses a weighted alternative to translate query terms and does not limit to the fixed translation from the dictionary as in DBQT.
For document translation, we find that both SMT and NMT have a similar performance which is close to PSQ.
The effectiveness of different approaches depends on the language pair (PSQ for EN-\textgreater{}SW and SMT for EN-\textgreater{}TL), which is a similar finding with \newcite{mccarley1999should} and \newcite{franz1999ad}.
In our experiments with deep relevance ranking models, we all use SMT and PSQ because they have strong performances in both language pairs and it is fair to compare.\\
\textbf{Effect of Extra Features and Bilingual Representation}.
While deep relevance ranking can achieve decent performance, the extra features are critical to achieve better results.
Because the extra features include the Indri score, the deep neural model essentially learns to rerank the document by effectively using a small number of training examples.
Furthermore, our models with bilingual representations achieve better results in both language pairs, giving additional 1-3 MAP improvements over their counterparts.
To compare language pairs, EN-\textgreater{}TL has larger improvements over EN-\textgreater{}SW.
This is because EN-\textgreater{}TL has better query translation, document translation, and query likelihood retrieval results from the baselines, and thus it enjoys more benefits from our model.
We also found POSIT-DRMM works better than the other two, suggesting term-gating is useful especially when the query translation can provide more alternatives.
We then perform ensembling of POSIT-DRMM to further improve the results. \\
\textbf{Zero-Shot Transfer Learning}.
Table \ref{tab:result_so} shows the result for a zero-shot transfer learning setting where we train on EN-\textgreater{}SW + EN-\textgreater{}TL and directly test on EN-\textgreater{}SO without using any Somali relevance labels.
This transfer learning delivers a 1-3 MAP improvement over PSQ and SMT.
This presents a promising approach to boost performance by utilizing relevance labels from other language pairs.

% \vspace{-2mm}
\section{Conclusion}
% \vspace{-2mm}
\label{sec:conclusion}
We propose to improve cross-lingual document retrieval by utilizing bilingual query-document interactions and learning to rerank from a small amount of training data for low-resource language pairs.
By aligning word embedding spaces for multiple languages, the model can be directly applied under a zero-shot transfer setting when no training data is available for another language pair.
We believe the idea of combining bilingual document representations using cross-lingual word embeddings can be generalized to other models as well.

\section*{Acknowledgements}
We thank Petra Galu\v{s}\v{c}\'{a}kov\'{a}, Douglas W. Oard, Efsun Kayi, Suraj Nair, Han-Chin Shing, and Joseph Barrow for their helpful discussion and feedback.
This research is based upon work supported in part by the Office of the Director of National Intelligence (ODNI), Intelligence Advanced Research Projects Activity (IARPA), via contract \# FA8650-17-C-9117. The views and conclusions contained herein are those of the authors and should not be interpreted as necessarily representing the official policies, either expressed or implied, of ODNI, IARPA, or the U.S. Government. The U.S. Government is authorized to reproduce and distribute reprints for governmental purposes notwithstanding any copyright annotation therein.

\bibliography{acl2019}
\bibliographystyle{acl_natbib}

\end{document}